\documentclass[letterpaper, 10 pt, conference]{ieeeconf}  

\IEEEoverridecommandlockouts                              

\usepackage{cite}
\usepackage{amsmath,amssymb,amsfonts}
\usepackage{algorithmic}
\usepackage{graphicx}
\usepackage{textcomp}
\usepackage{xcolor}

\usepackage{subcaption}
\usepackage{url}

\begin{document}

\title{Do We Run Large-scale Multi-Robot Systems on the Edge? \\More Evidence for Two-Phase Performance in System Size Scaling}

\author{Jonas Kuckling,\!$^{1,2}$\!
\and
Robin Luckey,\!$^{3}$\!
\and
Viktor Avrutin,\!$^{4}$\!
\and
Andrew Vardy,\!$^{5}$\!
\and
Andreagiovanni Reina,\!$^{2,6}$\!
\and
Heiko Hamann$^{1,2}$
\thanks{VA was supported by the German Research Foundation within project ``Generic bifurcation structures in piecewise-smooth maps with extremely high number of borders in theory and applications for power converter systems – 2''.
JK, AR, and HH acknowledge support from DFG through Germany’s Excellence Strategy-EXC 2117-422037984.
AR also acknowledges support from the Belgian F.R.S.-FNRS.}
\thanks{$^{1}$JK and HH are with the Department of Computer and Information Science, University of Konstanz, 78457 Konstanz, Germany.
        {\tt\footnotesize \{jonas.kuckling, heiko.hamann\}@uni-konstanz.de}.}%
\thanks{$^2$ 
JK, AR, and HH are also with the Centre for the Advanced Study of Collective Behaviour, University of Konstanz, Germany.}
\thanks{$^{3}$RL is with the Institute of Computer Engineering, University of Lübeck, Lübeck, Germany.}%
\thanks{$^{4}$VA is with the Institute for Systems Theory and Automatic Control, University of Stuttgart, Stuttgart, Germany.}%
\thanks{$^{5}$AV is with the Department of Computer Science and the Department of Electrical and Computer Engineering, Memorial University of Newfoundland, St. John's, Canada.}%
\thanks{$^{6}$AR is with IRIDIA, Université Libre de Bruxelles, Brussels, Belgium.}%
}

\maketitle

\begin{abstract}
With increasing numbers of mobile robots arriving in real-world applications, more robots coexist in the same space, interact, and possibly collaborate. 
Methods to provide such systems with system size scalability are known, for example, from swarm robotics. 
Example strategies are self-organizing behavior, a strict decentralized approach, and limiting the robot-robot communication. 
Despite applying such strategies, any multi-robot system breaks above a certain critical system size (i.e., number of robots) as too many robots share a resource (e.g., space, communication channel). 
We provide additional evidence based on simulations, that at these critical system sizes, the system performance separates into two phases: nearly optimal and minimal performance. 
We speculate that in real-world applications that are configured for optimal system size, the supposedly high-performing system may actually live on borrowed time as it is on a transient to breakdown. 
We provide two modeling options (based on queueing theory and a population model) that may help to support this reasoning.
\end{abstract}


\section{Introduction}

To scale up to larger and larger problems, robot tasks in industry, environmental/infrastructure monitoring, and our everyday lives will require larger and larger multi-robot systems, acting in parallel and cooperating~\cite{yang18,dorigo21}. 
With an increasing number of robots sharing resources, such as physical space and radio channels, scalability in system size will become an increasing challenge~\cite{hamann22}. 
Methods for designing robot systems that scale (at least for fixed swarm density) are developed and studied in swarm robotics~\cite{hamann18}. 
Even robots that were not designed to interact and cooperate may be required to do so in the future---further increasing the necessity to develop scalable systems.~\cite{rahwan19}.

Swarm robotics achieves scalability by enforcing a
decentralized approach based on self-organizing behaviors, local robot-robot interactions, and simple neighbor-based communication~\cite{hamann18}. 
Several studies have measured a certain intuitive swarm performance (e.g., number of collected items during a defined time interval in a foraging task~\cite{lu20}), usually over swarm density $\rho=N/A$, for swarm size~$N$ and a bounded operation area of size~$A$~\cite{kwa23,hunt19,khaluf17,hamann22}.
If area size~$A$ is fixed, scaling swarm size~$N$ effectively modulates the swarm density. 
Almost all studies that measure mean swarm performance over swarm density (e.g., see \cite{rosenfeld04}) show similar curves characterized by three stages: an initial increase with system size, reaching peak performance (at a critical system size~$N_c$), and for $N>N_c$ decreasing system performance. 
Interestingly, a diversity of systems from different fields of research show similar scalability results (e.g., see~\cite{ward93,mateo17,kerner09}). 
While measuring mean swarm performance is a good approach for an initial performance analysis, it is likely to hide an important property. 
The measured swarm performance may be bimodal showing two phases: an upper phase with high performance (i.e., each robot operates efficiently most of the time), a lower phase with minimal performance (i.e., most robots trapped in congestion etc.), and almost nothing in between~\cite{hamann13a}. 
This is, however, a potentially dangerous condition as swarm robotics systems often show linear or even super-linear performance increases for low swarm densities. 
By optimizing for performance, one may unknowingly put the system on the edge, where model predictions and real-world experience may catastrophically diverge. 
Also, the robot system likely has a long transient at critical system size with two implications. First, the system can actually be run on `borrowed time' if the transient is longer than what is relevant in applications (i.e., potentially contradicting mathematical predictions built on assumed convergence). Second, the system at runtime initially shows good performance but then suddenly breaks down completely.

As our main contribution, we show simulation results for a large real-world multi-robot system in a warehouse and two more, very different, academic and smaller scenarios (swarm robotics object clustering and emergent taxis) indicating clearly the two-phase situation. 
In addition, we provide two modeling options that are able to qualitatively describe this crucial two-phase system behavior. 
Which model may prove to be best for modeling two-phase swarm performance remains an open question at this point. 
It remains encouraging that different modeling techniques are able to represent the two-phase situation and yield quite similar predictions. 
Besides the obvious conclusion that a swarm engineer wants to avoid putting their robotic systems on the edge between performing maximally and breaking down completely, there are other immediate implications. 
For example, hysteresis effects are possible that may have extreme effects on the swarm performance at runtime. 
Known approaches of `robust scalability,' such as the online voluntary retreat of robots in order to reduce swarm density~\cite{mayya19} or of adapting algorithm parameters autonomously in reaction to a currently measured swarm density~\cite{wahby19d}, may trigger sudden performance jumps if the system operates close to a bifurcation point (see Sec.~\ref{sec:popModel}).

\section{Three Scenarios}
\subsection{Warehouse}
We consider an actual modern warehouse solution as required, for example, by today's online retailers of fashion to maximize economy of scale. 
The system is based on robots called `AFLE' (German abbreviation for Autonomous Driving Storage Units) that are currently under development by the company EMHS GmbH (see Fig.~\ref{fig:results:scenario:warehouse}). 
The idea is to build a large swarm (in the order of $10^5$) of low-cost mobile autonomous `hangers' as a solution to traditional pocket sorters instead of a dragging chain. 
Each AFLE carries a bag forming a storage unit. 
The robots hang on a rail grid that also provides power. 
By dividing the warehouse into independent segments the system has high potential to scale. 
To further enforce scalability, the system could combine central coordination by global messaging with decentralized approaches to coordinate pairs or small groups of AFLEs, for example, at crossings of rails. Hence, the system at least partially qualifies as a swarm robotics system. 

The simulation software is focused on large-scale simulations of warehouses with more than 3000 AFLEs. 
The warehouse is simplified to a grid world with grid cells that can have an edge per cardinal direction and contain not more than one robot.
All traffic is one-way and organized in `highways'; the actual storage rail sections are not modeled. 
Robots emerge on highways modeled by Poison distributions (i.e., mimicking the rather unpredictable influx of orders). 
The AFLE robots are routed with $A^*$. 
No physics is simulated but we penalize robots doing turns as the real robots would need to slow down.

\subsection{Object Clustering}

In the object clustering task a swarm of robots manipulate randomly placed objects, either bringing them to a pre-defined goal position, or aggregating them in a self-organized manner (see Fig.~\ref{fig:results:scenario:objectClustering}).  
If the goal is pre-defined, then this task becomes synonymous with foraging~\cite{zedadra2017multi}.  
We test the \emph{lasso method} for object clustering as its performance as a function of swarm size has already been partially explored~\cite{vardy2022lasso}.  
This method relies on an external scalar field which acts to guide the robots around the environment while nudging objects towards the global minimum of the field, located at the goal.  

The control algorithm operates independently on each robot. The most outlying visible puck is identified and the value of this puck's position on the scalar field is determined. 
A~control law drives the robot to reach and follow the contour line defined by this value.  
We observe that the robots tend to occupy the same contour line and travel clockwise around it, nudging pucks inwards toward the goal. 
Thus, a `lasso' is formed that tightens around the objects until they are forced to the goal position. Performance was found to increase with swarm size up to a point, then decrease due to spatial interference~\cite{vardy2022lasso}.
Fig.~\ref{fig:results:model:objectClustering} shows results collected in a custom simulator which models two-dimensional physics.\footnote{open-source code repository: \url{https://github.com/avardy/cwaggle_lasso}, video: \url{https://youtu.be/_KOU5SzpQBg}.}

\subsection{Emergent Taxis}

In the emergent taxis task \cite{hamann13a}, the objective is to move the swarm towards a light beacon  (see Fig.~\ref{fig:results:scenario:emergentTaxis}).
However, the robots only have an omnidirectional beacon sensor to identify the beacon.
That is they can only perceive if their line of sight to the beacon is unoccupied but no bearing information is available.
If a robot has a line of sight to the beacon, it is referred to as \emph{illuminated} otherwise it is considered \emph{shadowed}.
The robots move straight, until they approach a defined distance to another robot and trigger an obstacle avoidance behavior.
This \emph{avoid radius} depends on the state of the robot: if they are shadowed, the radius is smaller than if they are illuminated.
Thus, robots are biased to move towards the light beacon.
Additionally, a parameter $\alpha$ controls the coherence of the swarm.

With increasing swarm size, the effect of increased number of encounters influences the swarm performance.
Initially, the increased number of robots increases the density with which shadowed robots form behind and approach illuminated ones.
This in turn increases the number of times that an illuminated robot avoids a shadowed one and thus the movement bias towards the light beacon becomes stronger.
However, if the required coherence forces the robots to aggregate too tightly, the obstacle avoidance behavior might be continuously triggered thus slowing down any movement.
Previous results showed that performance increased with swarm size but a second phase of low performance emerged at higher swarm densities \cite{hamann13a}.

\section{Results}

Next, we present simulation results for each of the three scenarios. We focus on swarm performance over swarm size~$N$. Swarm performance is throughput for the warehouse scenario (i.e., how many autonomous hangers reach their destination within a given total experiment time), average distance between objects for the object clustering scenario, and the speed of the swarm's barycenter towards the light for the emergent taxis scenario. 
As we keep the operation area size~$A$ constant in all three cases, we effectively change the swarm density~$\rho$ by varying swarm size~$N$. In the case of the warehouse, we vary the swarm size~$N$ of coexisting robots indirectly via an arrival rate of how many robots are requested to approach the packing station per hour. 

\subsection{Warehouse}

\begin{figure*}
    \centering
    \subcaptionbox{\label{fig:results:scenario:warehouse}Warehouse prototype view}{\includegraphics[width=0.3\linewidth]{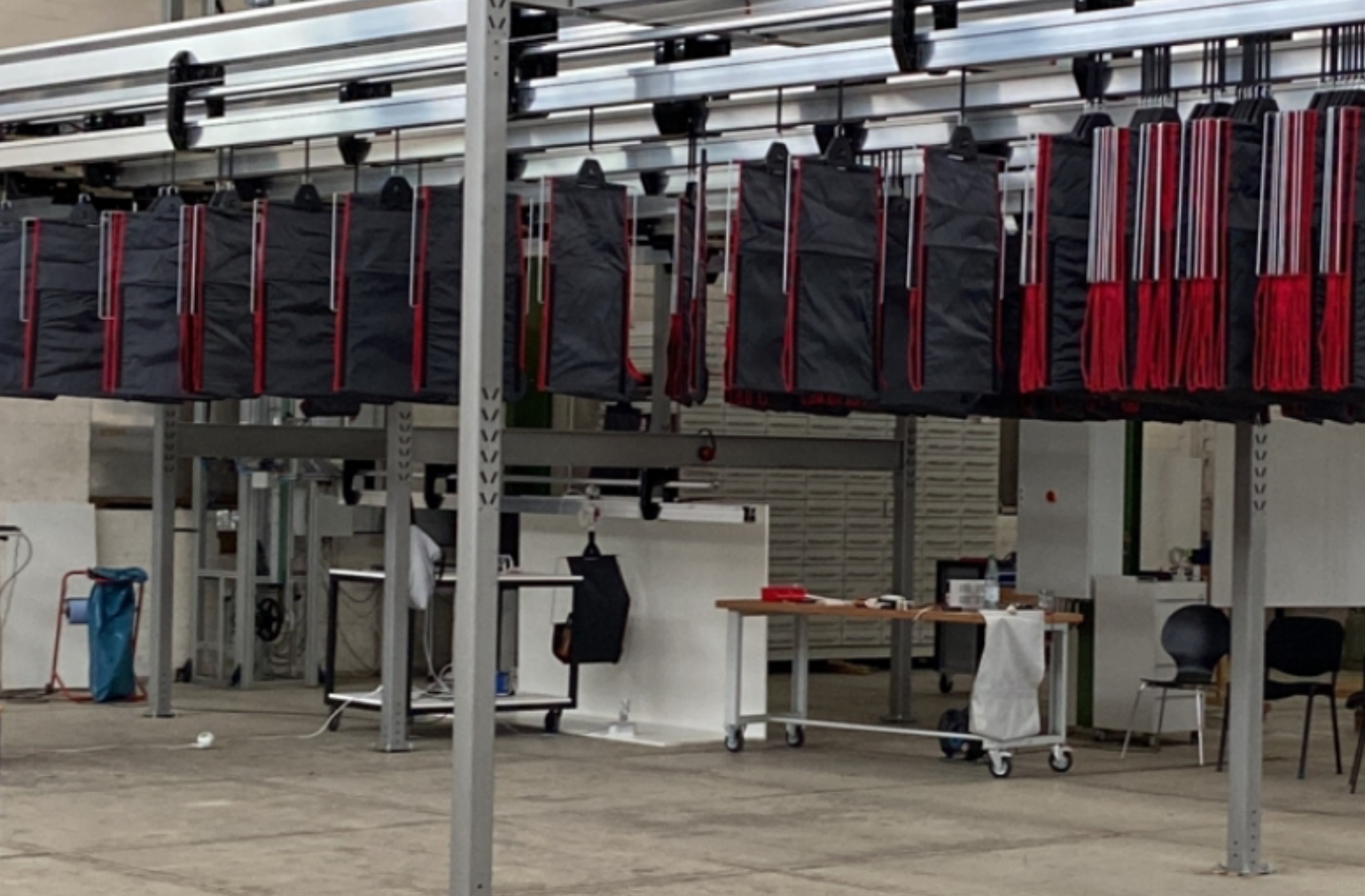}}%
    \hspace{1mm}
    \subcaptionbox{\label{fig:results:scenario:objectClustering}Object clustering overhead view}{\includegraphics[width=0.3\linewidth]{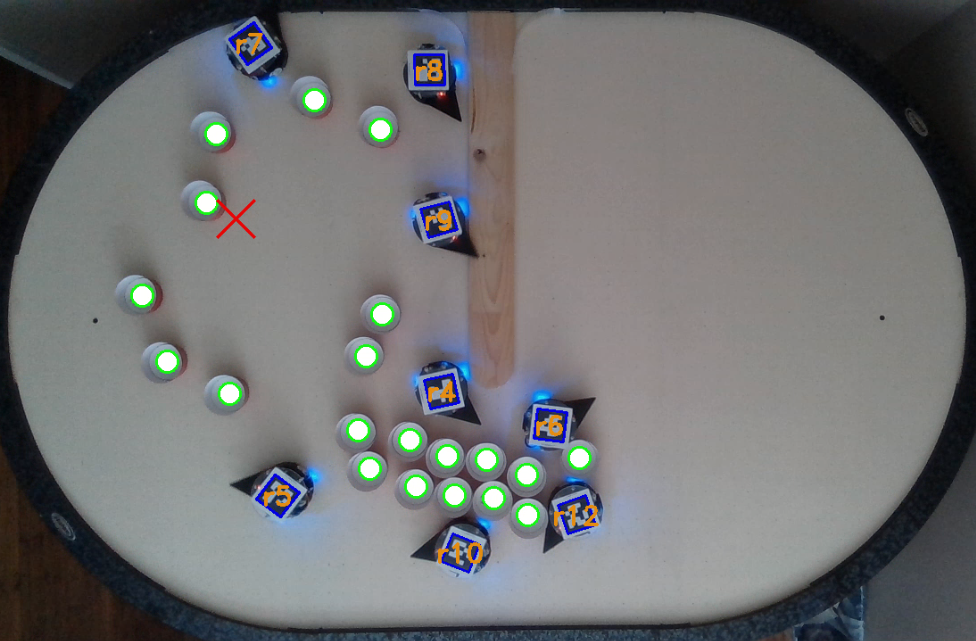}}%
    \subcaptionbox{\label{fig:results:scenario:emergentTaxis}Emergent taxis schematic}{\includegraphics[width=0.3\linewidth]{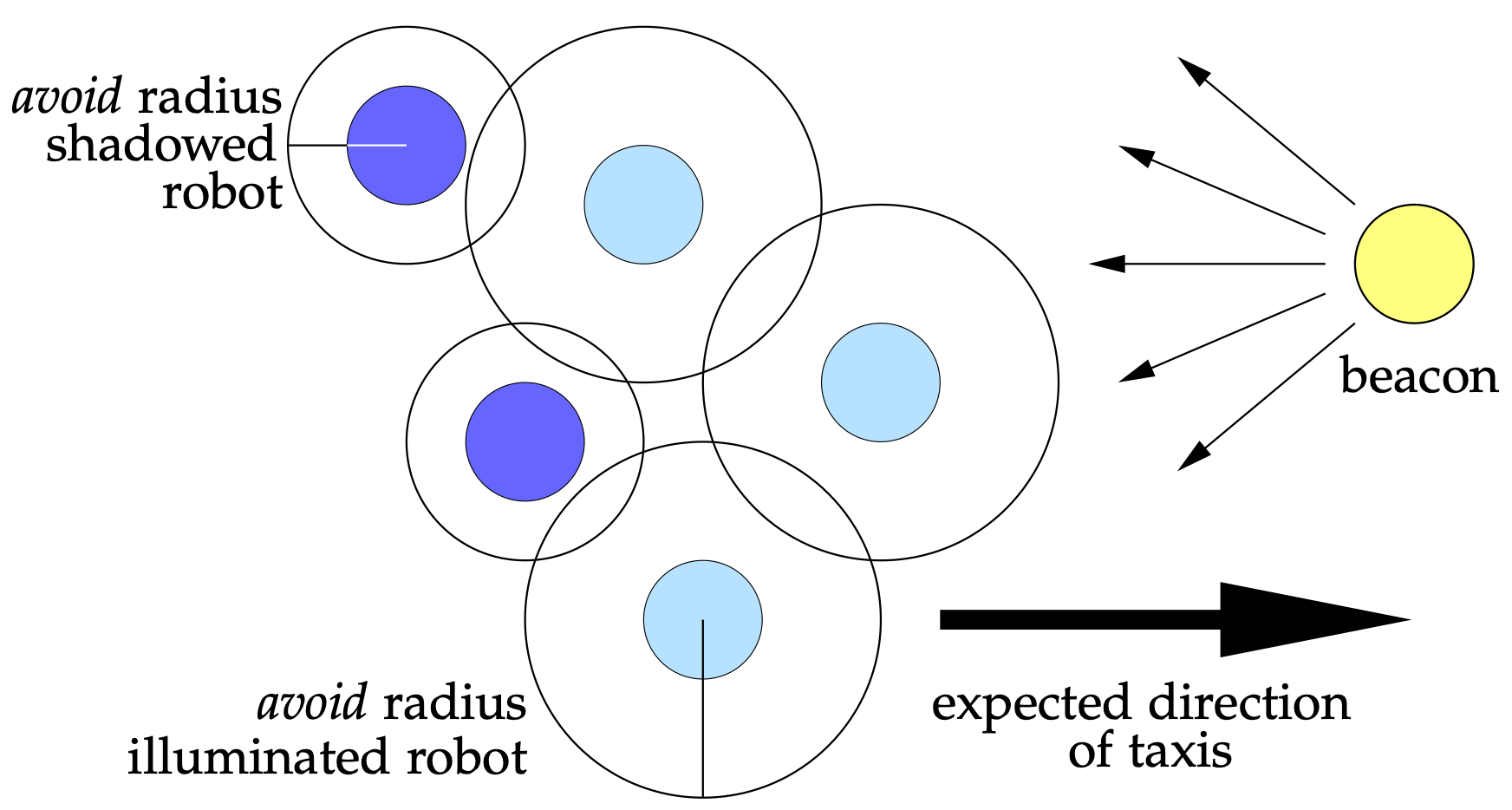}}%
    \\
    \subcaptionbox{\label{fig:results:model:wahrehouse}Warehouse results}{\includegraphics[width=0.3\linewidth]{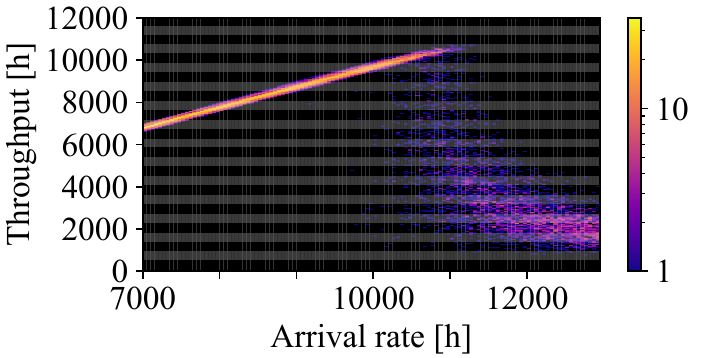}}%
    \subcaptionbox{\label{fig:results:model:objectClustering}Object clustering results}{\includegraphics[width=0.3\linewidth]{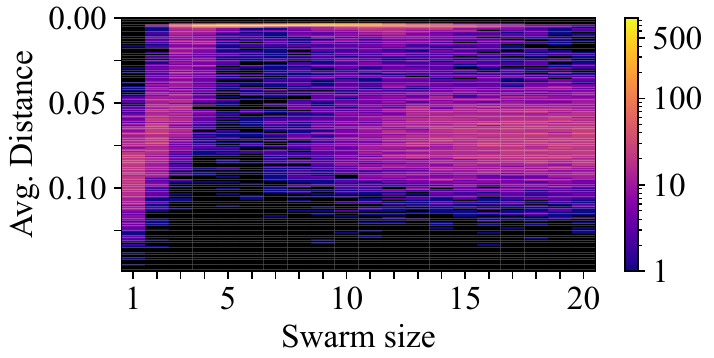}}%
    \subcaptionbox{\label{fig:results:model:emergentTaxis}Emergent taxis results}{\includegraphics[width=0.3\linewidth]{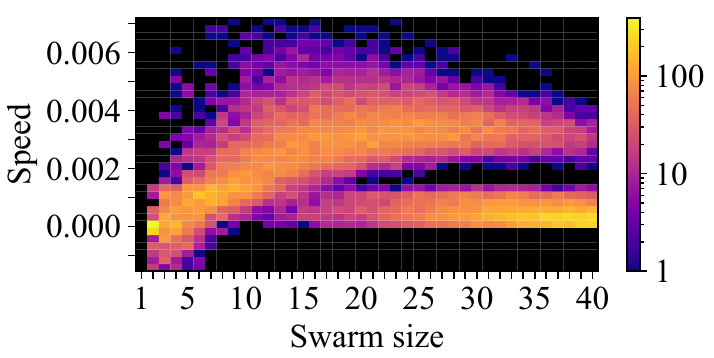}}%
    \caption{Upper row (a-c): Representative illustration of the three scenarios. Lower row (d-f): histograms of the swarm performance in simulation over swarm size (or arrival rate in the case of the warehouse scenario).}
    \label{fig:results}
\end{figure*}


Fig.~\ref{fig:results:model:wahrehouse} shows a 2d-histogram of the performance in simulation in the warehouse scenario.
We investigated different arrival rates with which new AFLEs are added to the warehouse.
The performance is measured as throughput, that is the number of AFLEs that reach their destination.
As AFLEs arrive and leave the warehouse during the experiment, the swarm size fluctuates over time.
However, assuming a constant time that the AFLEs spend inside the warehouse, the arrival rate can be an approximation for the number of AFLEs in the warehouse.
For each investigated value of the arrival rate, 96 repetitions were performed.
The color value of a cell in the histogram indicates how often a given arrival rate resulted in the corresponding throughput.

At low arrival rates, the system does not suffer from congestion and the throughput increases linearly with the arrival rate, as expected for a system with low density.
At an arrival rate of about $10^4$ units per hour,  interference effects may cause the throughput to drop considerably in few runs.
This is caused by a congestion in the warehouse, when too many AFLEs interfere with each other.
When two AFLEs are trying to traverse the same rail segment, they need to negotiate the traversal order and AFLEs with lower priority wait until AFLEs with higher priorities are passed.
With increased arrival rate, the number of experiments with congestion increases until all repetitions suffer from congestion.
Note that there is a range of arrival rate values for which part of the experiments operated properly, with maximum performance, and the remaining experiments collapsed into complete congestion with performance close to zero.
As a result, the distribution of the throughput splits into a high-performing and a low-performing \emph{phase}.

\subsection{Object Clustering}

\begin{figure}
    \centering
    \includegraphics[width=1.0\linewidth]{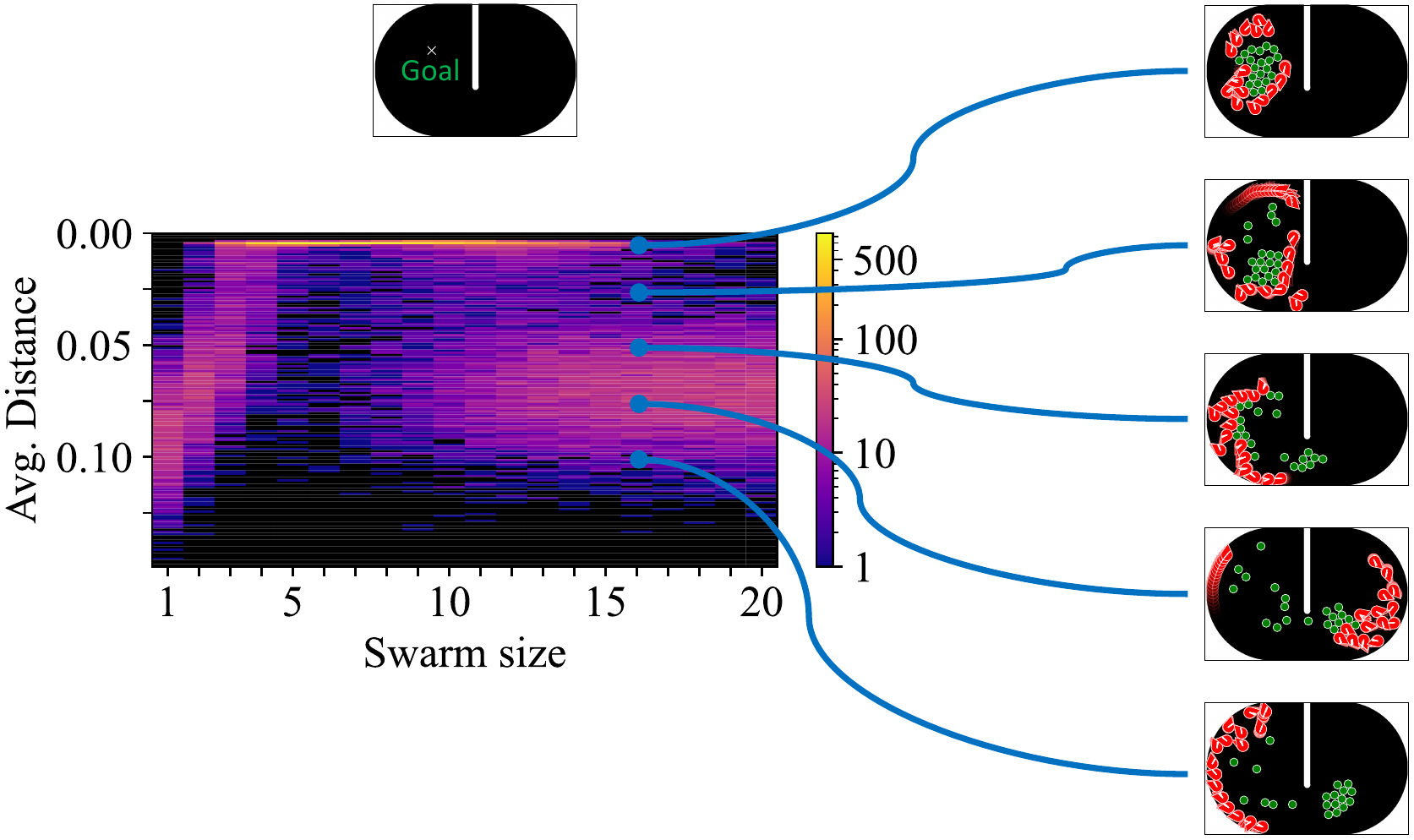}
    \caption{The performance plot from Fig.~\ref{fig:results:model:objectClustering} augmented with snapshots of the final arena configuration. 
    Robots are red. 
    Pucks are green.  The goal position is depicted at top-left.}
    \label{fig:result:clustering_performance_examples}
\end{figure}

Fig.~\ref{fig:results:model:objectClustering} shows the performance of simulated swarms performing the object clustering task, for swarm sizes from one to twenty robots.
For each swarm size, we evaluated the performance $1000$ times.
The performance is the normalized average distance of all objects to the target location. A~performance measure of $0$ is the ideal, but is not attainable due to physical interference between objects.

When the number of robots is too small, a small number of objects is nudged towards the goal, resulting in poor collective performance.
Performance increases with increasing swarm size, as predicted by theory with optimal performance around 6 robots.
For swarms with 10 or more robots, we observed that the performance splits into two phases---roughly, successful and unsuccessful.
Fig.~\ref{fig:result:clustering_performance_examples} provides examples of the final arena configuration with 16 robots for different levels of performance.  The top example shows a successful outcome in which all objects have been delivered to the goal.  The remaining examples represent situations where congestion has occurred and severely reduced performance.

\subsection{Emergent Taxis}


Fig.~\ref{fig:results:model:emergentTaxis} shows the performance in simulation of the swarm in the emergent taxis task. This result was published before and data was taken directly from the previous publication~\cite{hamann13a}. 
Each swarm size of $N\in\{2,3,\dots,40\}$ robots was evaluated 800 times.
At the beginning of the experiment, robots were distributed randomly on the far side of the arena.
The score of each evaluation was the speed with which the barycenter of the robot swarm moved towards the light during the experiment.
For negative speeds, the barycenter of the swarm moved away from the light.

For small swarm sizes, we did not observe any emergent taxis behavior.
However, with increasing swarm size, the swarm started moving towards the light and the speed of this taxis behavior also increased.
At around a swarm size of 15 robots, the maximum speed was reached and no further improvement was found.
At the same swarm size, we observed the emergence of a second phase, in which only the observed speed of the taxis behavior was low or sometimes nonexistent.
In the first phase, the swarm is still able to function properly, whereas in the second case it becomes \emph{pinned}, with many obstacle avoidance maneuvers triggered resulting in only slow movements.

\section{Models}

Next, we propose two modeling approaches that capture the two-phase character of swarm performance as observed in real-world systems. The first is a simple method based on queueing theory as an example of low model complexity from computer science that is still capable of describing key features. 
As second method we use our previously published population model~\cite{hamann22} that we introduced as a generic approach to model multi-robot systems but also large parallel computing systems. It comes with a lot of complexity but also proves to be capable of catching key properties of the swarm systems including predictions for transient times, that is for example, how long a real-world robot system would perform high despite sitting on the edge. We provide a full discussion of the resulting nonlinear system in terms of nonlinear dynamics. We believe that especially the model predictions of transient times may prove to be relevant for real-world applications. Hence, a deep understanding of all the mathematical implications is essential for a complete (future) understanding of scalability in robot swarms. 

\subsection{Queues}
\begin{figure}
    \centering
    \subcaptionbox{\label{fig:model:mm1:a}throughput and queue length~$N_q$}{\includegraphics[width=0.8\linewidth]{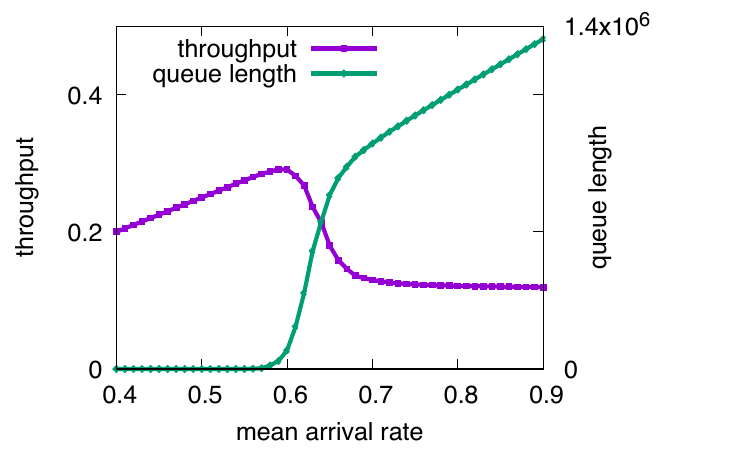}}%
    \\
    \subcaptionbox{\label{fig:model:mm1:b}throughput histogram (frequency color-code log.)}{\includegraphics[width=0.8\linewidth]{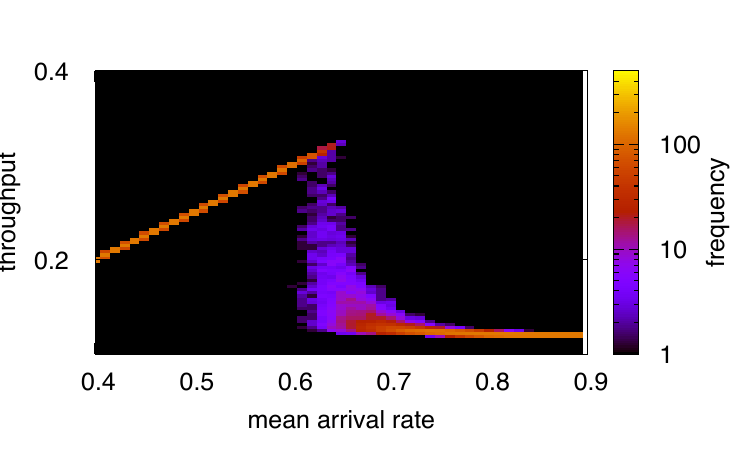}}%
    \caption{Queuing model: throughput and queue length~$N_q$ over arrival rate for modified M/M/1 queue (service time dependent on queue length).}
    \label{fig:model:mm1}
\end{figure}

A simple approach of modeling the two-phase situation of swarm performance can be based on queueing theory. 
An intuitive and sophisticated model could be based on queueing networks to explicitly represent the shared resources, such as space and radio bandwidth (possibly also truncated queues). 
However, a simpler approach is possible by an unorthodox change of the standard M/M/1 queue with exponential interarrival-time distribution $\lambda\exp{(-\lambda t)}$, exponential service time $\mu\exp{(-\mu t)}$, one channel, infinite capacity. 
Key change to the standard model is that we make the service time dependent on the current queue length~$N_q(t)$ (i.e., a cost on keeping the queue) by setting mean service time~$1/\mu(t)=1/(c_1\log(c_2N_q(t))+\mu_c)$ for arbitrary constants~$c_1=0.3$ and $c_2=0.2$. For the following example we set $\mu_c=1$. 
We simulate the queue for mean arrival rates~$\lambda\in[0.4,0.9]$ (500 samples each) for $2\times 10^6$ time steps and exclude data from the first $1\times 10^6$ time steps as transient. 
Resulting mean throughput and mean queue length are shown in Fig.~\ref{fig:model:mm1:a} and a histogram of the throughput in Fig.~\ref{fig:model:mm1:b}. 
It is to note that the curves shown in Fig.~\ref{fig:model:mm1:a} resemble those predicted by the population model (see Fig.~\ref{fig:fixed:points}a).
This queuing model is able to represent the linear increase and the breakdown of the system at a critical arrival rate~$\mu_c\approx 0.65$. 

\subsection{Population model and nonlinear dynamics analysis}
\label{sec:popModel}

\begin{figure}
    \centering
\includegraphics[width=0.7\linewidth]{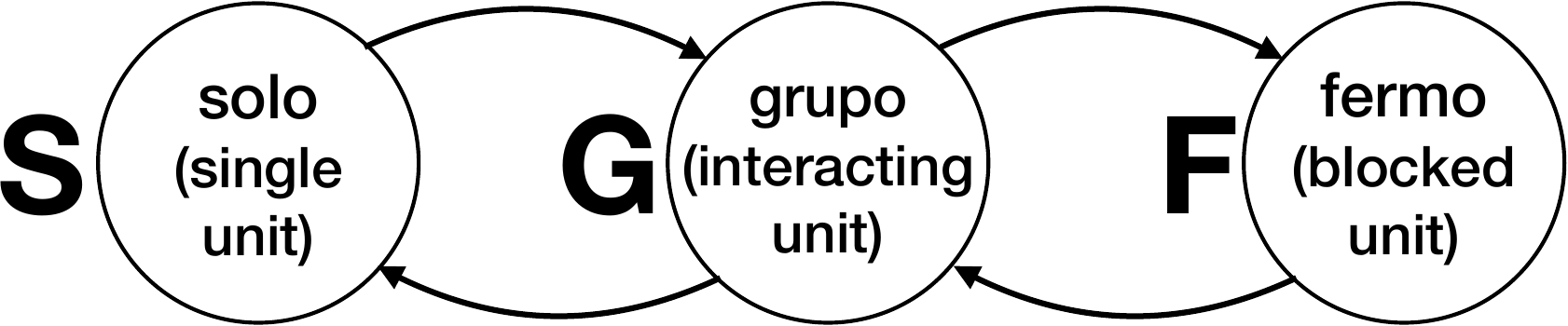}%
    \caption{Underlying schema for population model: transitions for robots in state \emph{solo}, \emph{grupo}, and \emph{fermo}~\cite{hamann22}.}
    \label{fig:model:schema}
\end{figure}

In previous work, we have presented a population model that grasps most of the relevant system features concerning scalability~\cite{hamann22}. 
Key advantages are that it is a general, unified model that abstracts and describes scalability patterns across different systems, including parallel supercomputers, robot swarms, and wireless sensor networks, based on the microscopic interactions between units. 
In this alternative approach to model scalability we assume that each robot can be assigned to one of three abstract robot states relevant for scalability analysis. The system dynamics is based on $N$~robots that transit between these three operational states. We assume robots can be in state \emph{solo} (operating alone without sharing communication and/or space resources with others), in state \emph{grupo} (sharing resources with other robots), or in state \emph{fermo} (interfering with others in the use of resources). The model describes through seven transition rates ($k_1,\dots,k_7$) the probability that robots change their state as a consequence of independent actions or interactions with other robots. 
As in any population model we make assumptions, for example, that it is more likely for a \emph{solo} robot to approach a \emph{solo} robot if the \emph{solo} state currently has a high percentage in the swarm. We assume also that two \emph{solo} robots approaching each other trigger transitions to \emph{grupo} robots and similar for any other interaction of two robots (see Fig.~\ref{fig:model:schema}). 
%
The model is expressed as a system of ordinary differential equations (ODEs):
\begin{equation}
    \left\{
    \begin{aligned}
      \frac{ds}{dt} &=-2k_1 s^2-k_2 s g - k_3 s f + k_4 g \\
      \frac{df}{dt} &= 2k_5 g^2 + k_6 g f - k_7 f \,,
    \end{aligned}
    \right.
    \label{eq:2ODE}
\end{equation}
with $g=N-s-f$ due to the robot conservation assumption.
The system variables $s$, $g$, and $f$ represent the proportion of
robots in states \emph{solo}, \emph{grupo}, and \emph{fermo}, respectively. 
$(s^*,g^*,f^*)$, the stable fixed point of system \eqref{eq:2ODE}, describes the long-term distribution of the robots among the three states. 

\begin{figure*}
\centering
  \subcaptionbox{}{\includegraphics[width=.32\textwidth]{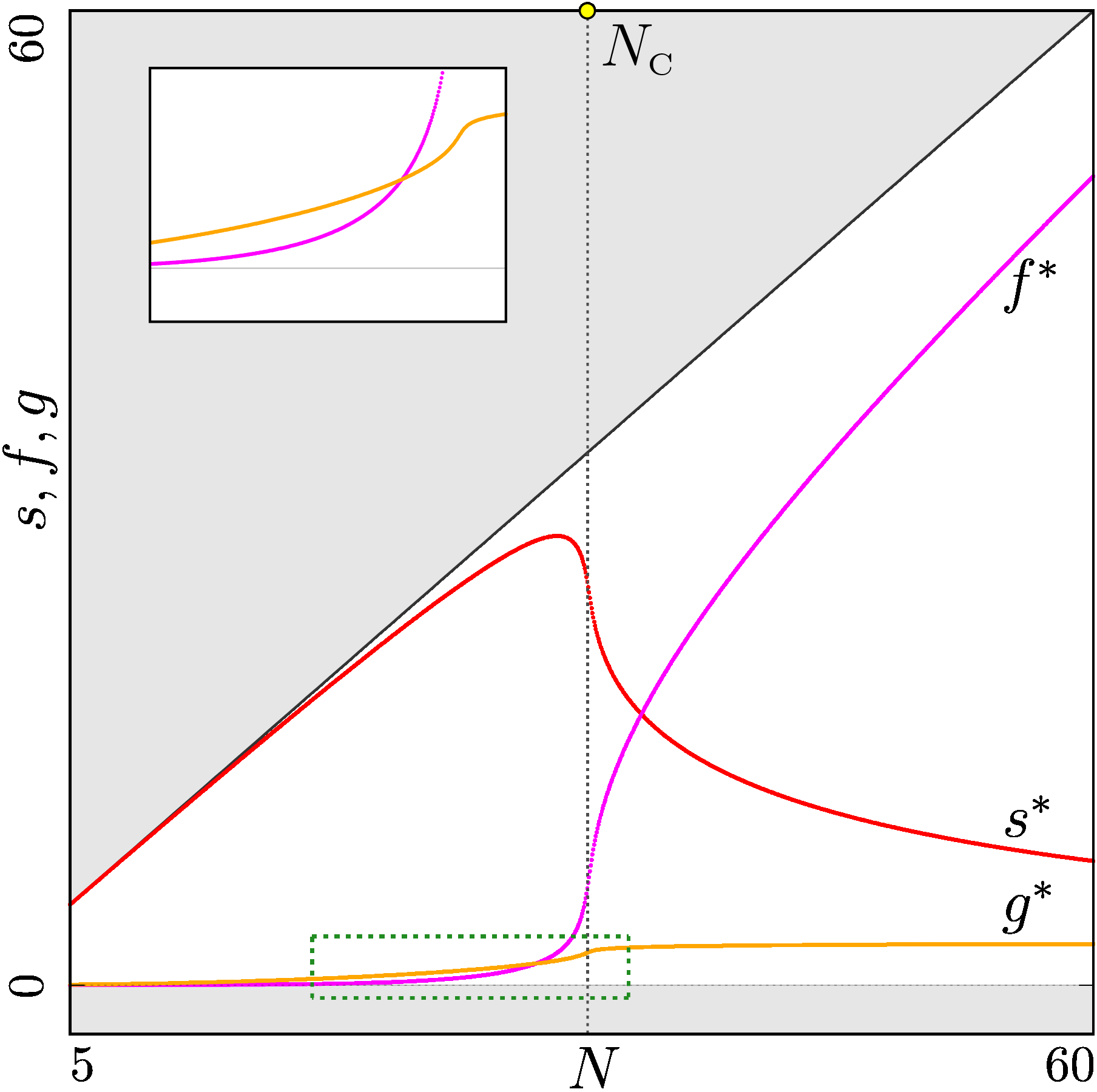}}
  \hfill
  \subcaptionbox{}{\includegraphics[width=.32\textwidth]{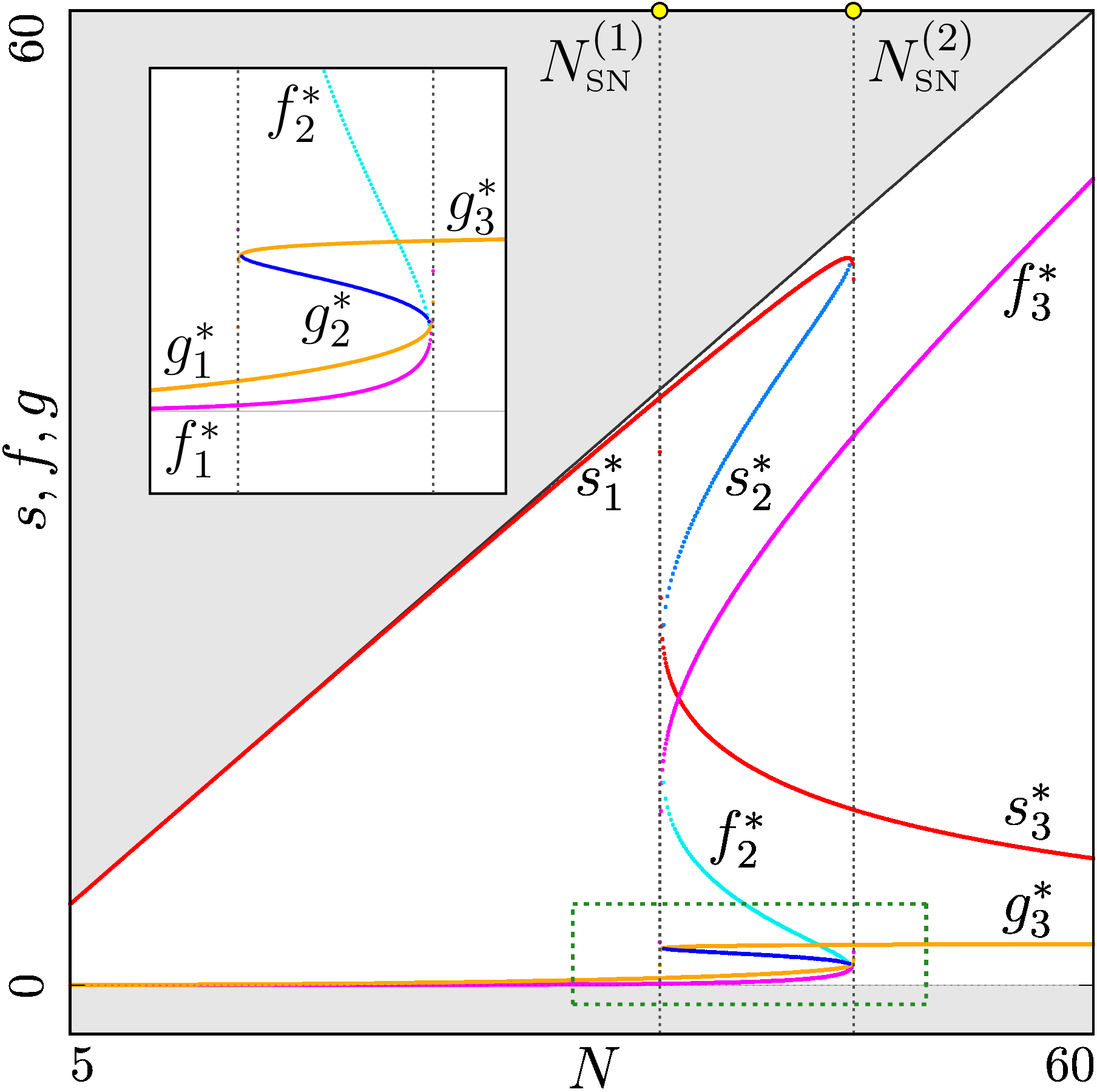}}
  \hfill
  \subcaptionbox{}{\includegraphics[height=.32\textwidth]{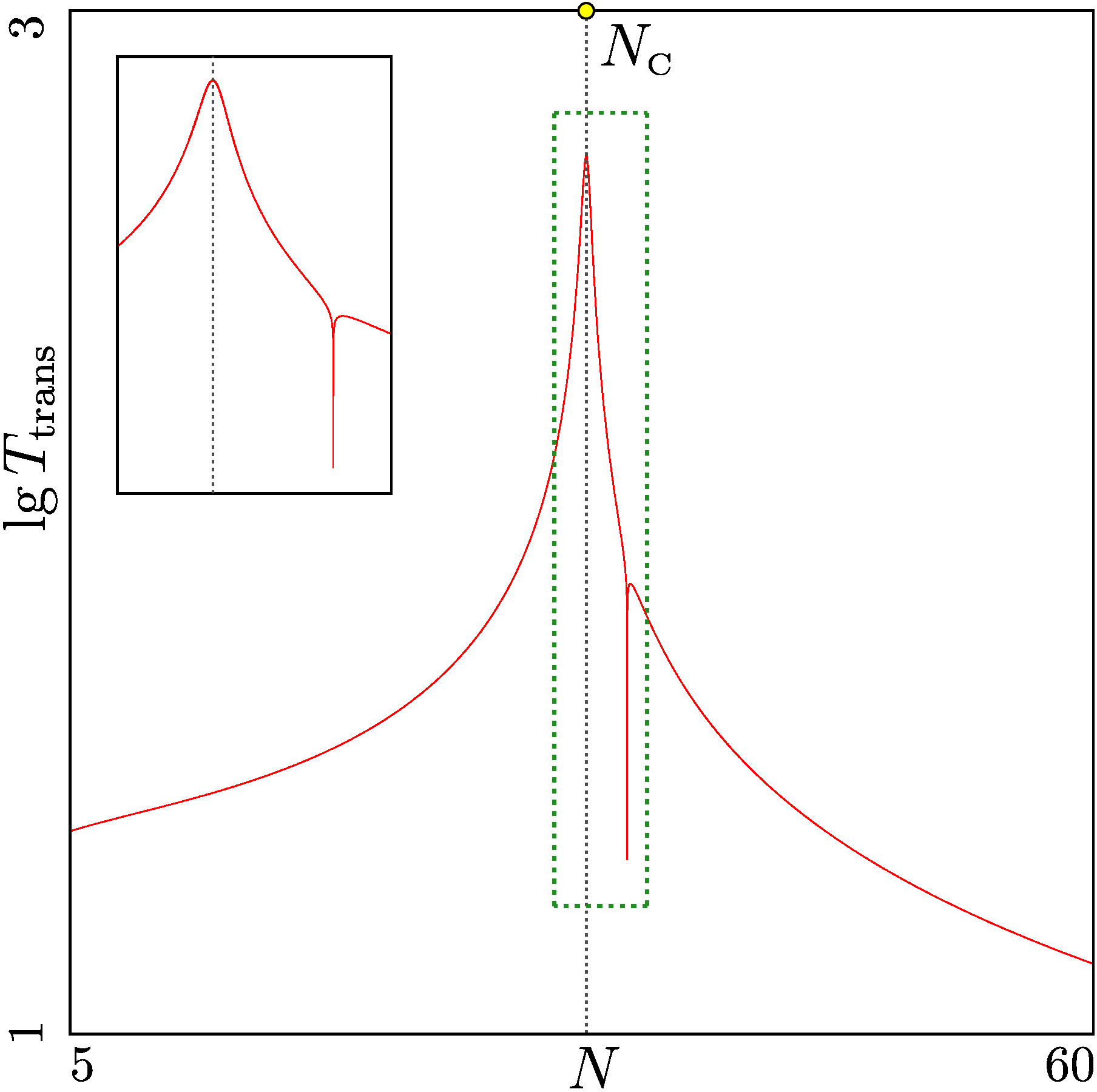}}
  \caption{Stationary states of Eq.~\eqref{eq:2ODE}
    under variation of $N$
    for (a) $k_1=0.005$ and (b)
    $k_1=0.001$, other parameters as in Eq.~\eqref{eq:parameter:values}.
    (b)~Domain between the saddle-node bifurcation
    points $N^{(1)}_{\text{\tiny SN}}$, $N^{(2)}_{\text{\tiny SN}}$
    associated with bi-stability,
    the non-physical domain outside of $0\leq s,f,g \leq N$
    is shown in gray;
    (c)~transient time $T_{\text{trans}}$ for fixed 
    initial state $(s_0,f_0)=(0,0)$ and convergence to stationary 
    state $(s_1^*, f_1^*)$ with accuracy $10^{-12}$;
    $k_1=0.005$. Insets show the indicated rectangles magnified.
  }
  \label{fig:fixed:points}
\end{figure*}

\begin{figure}\centering
  \includegraphics[height=.33\textwidth]{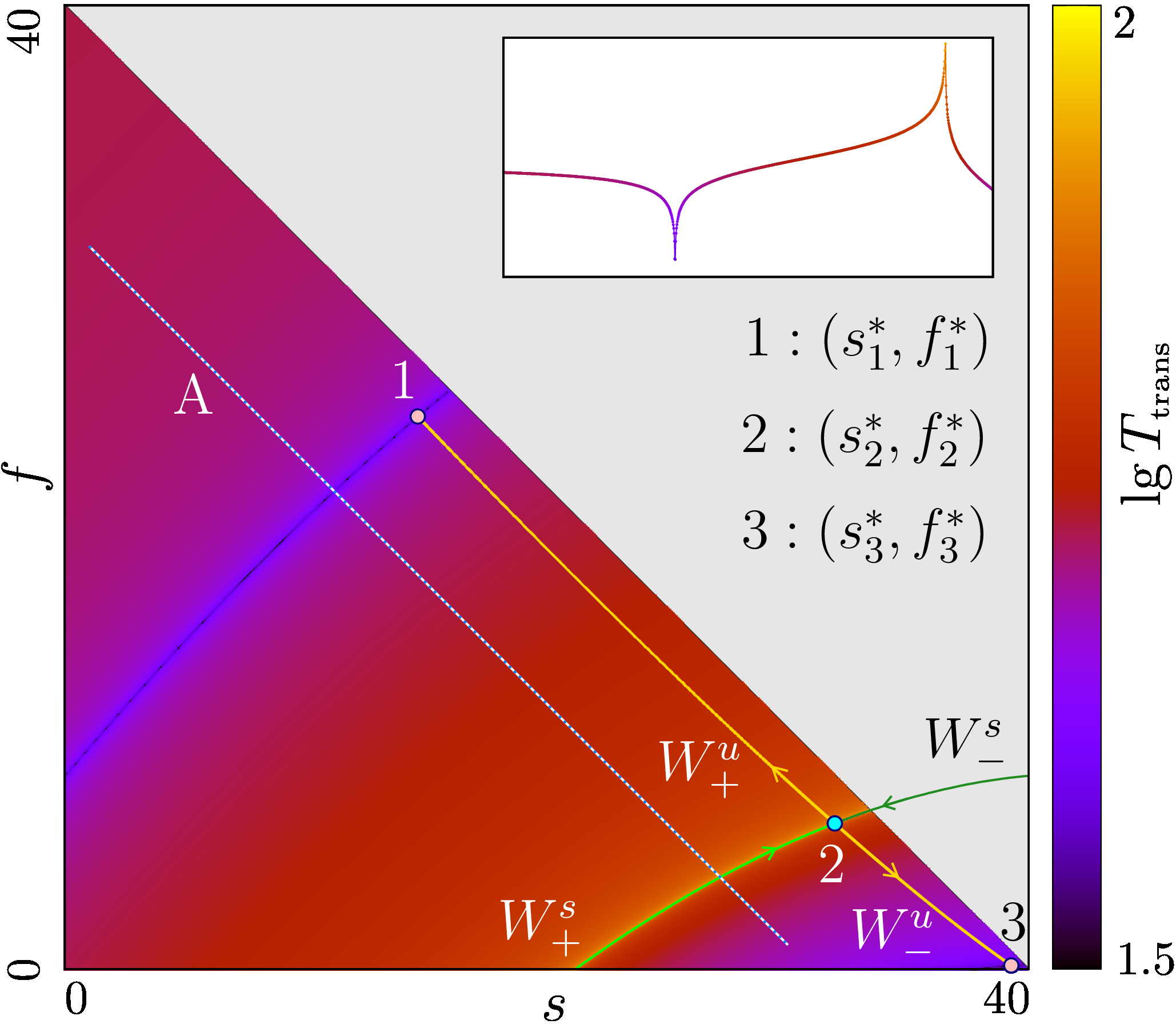}  
  \caption{\emph{fermo}-\emph{solo} state space in the bi-stable domain ($N=40$); color-coded logarithmic transient time $T_{\text{trans}}$, that is, convergence towards stable stationary states $(s_1^*, f_1^*)$ and $(s_3^*, f_3^*)$ with the same accuracy of $10^{-12}$. 
    Also the saddle point $(s_2^*, f_2^*)$ and its stable and unstable manifolds ($W^s_{\pm}$ and $W^u_{\pm}$, respectively) are shown.
    The inset shows the transient times for initial values located at the path indicated with A.
    }
  \label{fig:basins}
\end{figure}

In the numerical simulations presented below, we use the following set of parameters, unless stated otherwise:
\begin{equation}
\label{eq:parameter:values}
\begin{split}
&
    k_1=0.005,\:
    k_2=0.1,\:
    k_3=0.06,\:
    k_4=10,\\
&
    k_5=0.15,\:
    k_6=0.3,\:
    k_7=0.8
\end{split}            
\end{equation}

For a quick understanding of this model, we have analyzed the stationary system behavior. 
Fig.~\ref{fig:fixed:points}a shows the stationary states $(s^*,f^*)$ of
system~\eqref{eq:2ODE} at the parameter set specified 
in Eq.~\eqref{eq:parameter:values} under
variation of swarm size~$N$. 
For completeness, the resulting value $g^*=N-s^*-f^*$ due to robot conservation is shown, too. 
Fig.~\ref{fig:fixed:points}a shows, how for small swarm sizes~$N$, the number of \emph{solo} robots~$s^*$ grows almost linearly (most robots remain in \emph{solo}, few in \emph{fermo} and \emph{grupo}). 
%
For increasing swarm size~$N$, a sudden change of
the system's configuration occurs. The number of \emph{solo} robots reaches a
maximum (here, $N=31$) and then 
decrease rapidly. 
Simultaneously, we get more \emph{fermo} robots,
resulting eventually in their predominance. 
From a nonlinear dynamics perspective, the shifted 
location of the asymptotically stable fixed point is not a
bifurcation as the topological structure of
the state space remains unchanged under parameter variation. 
However, in practical applications the sudden decrease of \emph{solo} robots would have a significant impact on system performance. 

For other parameter values we may observe more sophisticated 
dynamics. As an example, Fig.~\ref{fig:fixed:points}b shows the
stationary states of the system for the value $k_1=0.001$ instead of
$k_1=0.005$ (i.e., slower transition of \emph{solo} robots to \emph{grupo} robots). As before, we
observe an almost linear growth of \emph{solo} robots until it reaches a maximal value.
However, soon after, the stable solution $(s_1^*,f_1^*)$ merges with 
an unstable one
$(s_2^*,f_2^*)$ and disappears in a saddle-node bifurcation at $N=N^{(2)}_{\text{\tiny
    SN}}$.  The unstable stationary state $(s_2^*,f_2^*)$ involved in this bifurcation
appears already before, via a different saddle-node bifurcation which
occurs at $N=N^{(1)}_{\text{\tiny SN}} < N^{(2)}_{\text{\tiny SN}}$
and gives rise to another asymptotically stable stationary
state $(s_3^*,f_3^*)$. Thus, in the parameter interval $N^{(1)}_{\text{\tiny
    SN}}<N<N^{(2)}_{\text{\tiny SN}}$, two stable stationary states
$(s_1^*,f_1^*)$, $(s_3^*,f_3^*)$ coexist and the system exhibits 
a bi-stable behavior, so that
 it may converge either to the stationary state where
\emph{solo} robots prevail, or to the one characterized by the prevalence of
\emph{fermo} robots, depending on initial values~$s_0$ and~$f_0$.

In Fig.~\ref{fig:basins}, we illustrate the system's state space. 
As two saddle-node bifurcations occurring at $N=N^{(1)}_{\text{\tiny SN}}$ and $N=N^{(2)}_{\text{\tiny SN}}$ are connected by the unstable (saddle)
point $(s_2^*,f_2^*)$, the basins of attraction of the stable stationary states are
separated from each other by the branches $W^s_{\pm}$ of the stable manifold of
the saddle $(s_2^*,f_2^*)$.
Here, the initial values located above or below $W^s_{\pm}$ 
converge towards $(s_1^*,f_1^*)$ or $(s_3^*,f_3^*)$, respectively.

Importantly for any application,  the bi-stable dynamics lead necessarily to a hysteresis effect. If we start from small swarm sizes~$N$ and increase swarm size incrementally, then the system follows the upper branch of the solutions, that is, the stationary state $(s_1^*,f_1^*)$ (see Fig.~\ref{fig:fixed:points}b). 
At the bifurcation point $N=N^{(2)}_{\text{\tiny SN}}$, the system undergoes a sudden jump to the lower branch $(s_3^*,f_3^*)$ (the saddle-node bifurcation acts as a so-called hard bifurcation, or a hard loss of stability~\cite{Arnold2003}). If one additional robot is introduced ($N+1$), the system undergoes a sudden change, which corresponds to a sudden catastrophic loss of productivity.
This roughly resembles the Braess paradox from game theory~\cite{Braess1968} where the overall performance of a system may decrease as an additional resource is provided.
Instead, if we start from a large swarm size~$N$ and we switch off robots one by one, then the system follows the lower branch of solutions (stationary state $(s_3^*,f_3^*)$) operating at low productivity until the bifurcation point~$N=N^{(1)}_{\text{\tiny SN}}$. Then, the system jumps to the upper branch $(s_1^*,f_1^*)$ and the productivity is suddenly increased. 
%


Important for all applications is the transient time (i.e., time to converge---with a given accuracy---to the stationary state).
As an example, Fig.~\ref{fig:fixed:points}c shows the transient time on logarithmic scale corresponding to the bifurcation diagram in Fig.~\ref{fig:fixed:points}a.
The transient time grows over-exponentially as the system approaches the critical state~$N_{\text{\tiny C}}$. This behavior, known as \emph{critical slowing down}, is typically observed in a neighborhood of a bifurcation point. This may be surprising, since, as already mentioned, there is no bifurcation (in the rigorous sense) in Fig.~\ref{fig:fixed:points}a. However, in Fig.~\ref{fig:fixed:points}b (different parameter~$k_1$) we observe two saddle-node bifurcations. Therefore, in the 2D parameter space~$(N,k_1)$ there exists a codimension-2 \emph{cusp}-bifurcation point~\cite{Kuznetsov04} from which two saddle-node bifurcation points emerge, leading to bistability and hysteresis. Hence, the long transient time close to the critical state (see Fig.~\ref{fig:fixed:points}a) is indeed caused by the critical slowing down in a neighborhood of a cusp bifurcation in 2D parameter space even if this bifurcation is not visible in the bifurcation diagram shown in Fig.~\ref{fig:fixed:points}a. 
%
In addition, the transient time $T_{\text{trans}}$ depends not exclusively on the parameters but also on the initial values~$s_0$, and~$f_0$. The transient time grows (over-exponentially, tending to infinity) for initial values close to the boundary between the basins of attraction of the coexisting stable fixed points (see Fig.~\ref{fig:basins}).

%

\section{Discussion and Conclusion}

We have presented three multi-robot systems that show the characteristic two-phase performance signature where, for a critical swarm size, the phases of high and low performance coexist. We have presented simulations of the AFLE robot system by EMHS GmbH, showing that not only robot swarms but also real-world applications of actual productive systems in existing industries show this same property in their system size scaling. 
We have also presented simulation results of an object clustering scenario that was previously implemented on real robots, showing the validity of the simulation~\cite{vardy2022lasso}. Also in this more academic example, we observe the same two-phase performance signature. 
As a last example, we have reminded the reader of simulation results for the well-known emergent taxis scenario that were published before~\cite{hamann13a}. This original finding inspired the presented research. 
To point to a possible interpretation of these results, we have presented two modeling options based on queueing theory and population models. 
The adapted M/M/1~queue is able to catch most of the discussed scalability features despite its simplicity. Our key contribution in terms of modeling, however, is the population model and especially its interpretation based on nonlinear dynamics. 
These interpretations point to three important findings and issues. (a)~There is indeed a bifurcation in the system and rapid changes in performance with small changes in swarm size should be expected. (b)~In the vicinity of bifurcation points we should indeed expect extremely increased transient times. (c)~We should expect hysteresis effects depending on swarm size that is increased or decreased online while operating. 
Swarm robotics systems that are optimized for performance may unknowingly be put on the edge with the effect of unexpected, catastrophic performance breakdowns. 
Long transients may either obscure this effect or may buy the robot swarm time that is possibly long enough to survive in high-performing states for the whole mission time. 
Future research will hopefully provide more evidence of whether the two-phase performance signature in scalability of swarm robotics is real and relevant even in industrial applications.

\section*{Acknowledgment}
HH thanks Pawel Romanczuk for the valuable discussions.

\bibliographystyle{IEEEtran} 
\bibliography{IEEEabrv,refs}

\end{document}